\documentclass[final]{cvpr}

\usepackage{times}
\usepackage{epsfig}
\usepackage{graphicx}
\usepackage{amsmath}
\usepackage{amssymb}
\usepackage{amsthm}
\usepackage{url}
\usepackage{algorithm}
\usepackage{algorithmic}
\usepackage{booktabs,multirow}
\usepackage[misc]{ifsym}

\usepackage[pagebackref=true,breaklinks=true,colorlinks,bookmarks=false]{hyperref}

\def\cvprPaperID{656} % *** Enter the CVPR Paper ID here
\def\confYear{CVPR 2021}

\newtheorem{prop}{Condition}
\newtheorem{lemma}{Lemma}
\newtheorem{cor}{Corollary}

\newcommand{\e}[1]{\ensuremath{\times 10^{#1}}}

\newcommand*{\affaddr}[1]{#1} % No op here. Customize it for different styles.
\newcommand*{\affmark}[1][*]{\textsuperscript{#1}}

\begin{document}

%%%%%%%%% TITLE
\title{Training Networks in Null Space of Feature Covariance for Continual Learning \thanks{Accepted as an oral of CVPR 2021}}

%\author{Shipeng Wang, Xiaorong Li, Jian Sun\thanks{corresponding author}, Zongben Xu\\
%School of Mathematics and Statistics, Xi'an Jiaotong University\\
%Xi'an, 710049, China\\
%{\tt\small \{wangshipeng8128, lixiaorong\}@stu.xjtu.edu.cn, \{jiansun, zbxu\}@xjtu.edu.cn }
%}
\author{
	Shipeng Wang\affmark[1], Xiaorong Li\affmark[1], Jian Sun${(\textrm{\Letter})}$\affmark[1,2,3], Zongben Xu\affmark[1,2,3]\\
	\affaddr{\affmark[1] School of Mathematics and Statistics, Xi'an Jiaotong University, Xi'an, 710049, China}\\
	\affaddr{\affmark[2] National Engineering Laboratory of
	Big Data Algorithms and Analysis Technology, Xi'an, 710049, China}\\
	\affaddr{\affmark[3] Pazhou Lab, Guangzhou, Guangdong, 510335, China}\\
	{\tt\small \{wangshipeng8128, lixiaorong\}@stu.xjtu.edu.cn, \{jiansun, zbxu\}@xjtu.edu.cn }
}
\maketitle
\thispagestyle{empty}
\pagestyle{empty}

%%%%%%%%% ABSTRACT
\begin{abstract}
	In the setting of continual learning, a network is trained on a sequence of tasks, and suffers from catastrophic forgetting. To balance plasticity and stability of network in continual learning, in this paper, we propose a novel network training algorithm called Adam-NSCL, which sequentially optimizes network parameters in the null space of previous tasks. We first propose two mathematical conditions respectively for achieving network stability and plasticity in continual learning. Based on them, the network training for sequential tasks can be simply achieved by projecting the candidate parameter update into the approximate null space of all previous tasks in the network training process, where the candidate parameter update can be generated by Adam. The approximate null space can be derived by applying singular value decomposition to the uncentered covariance matrix of all input features of previous tasks for each linear layer. For efficiency, the uncentered covariance matrix can be incrementally computed after learning each task. We also empirically verify the rationality of the approximate null space at each linear layer. We apply our approach to training networks for continual learning on benchmark datasets of CIFAR-100 and TinyImageNet, and the results suggest that the proposed approach outperforms or matches the state-ot-the-art continual learning approaches. 
\end{abstract}
%along with the batch-based network optimization
% batch-wisely

%%%%%%%%% BODY TEXT
\section{Introduction}

Deep neural networks have  achieved promising performance on various tasks in natural language processing, machine intelligence, etc., \cite{bahdanau2014neural,devlin2018bert, silver2016mastering,spampinato2017deep,sun2016deep}. However, the ability of deep neural networks for continual learning is limited, where the network is expected to continually learn knowledge from sequential tasks \cite{Hsu18_EvalCL}. The main challenge for continual learning is how to overcome catastrophic forgetting \cite{french1999catastrophic,mccloskey1989catastrophic,ratcliff1990connectionist}, which has drawn much attention recently. 

In the context of continual learning, a network is trained on a stream of tasks sequentially. The network is required to have \textit{plasticity} to learn new knowledge from current task, and also \textit{stability} to retain its performance on previous tasks. However, it is challenging to simultaneously achieve plasticity and stability in continual learning for deep networks, and catastrophic forgetting always occurs. This phenomenon is called \textit{plasticity-stability dilemma} \cite{mermillod2013stability}. %where plasticity is the ability to learn new knowledge from current task while stability is the ability fo retain the performance.

Recently, various strategies for continual learning have been explored, including regularization-based, distillation-based, architecture-based, replay-based and algorithm-based strategies. The regularization-based strategy focuses on penalizing the variation of parameters across tasks, such as EWC \cite{kirkpatrick2017overcoming}. The distillation-based strategy is inspired by knowledge distillation, such as LwF \cite{li2017learning}. The architecture-based strategy modifies the architecture of network on different tasks, such as \cite{abati2020conditional,li2019learn}. The replay-based strategy utilizes  data from previous tasks or pseudo-data to maintain the network performance on previous tasks, such as \cite{aljundi2019gradient,ostapenko2019learning}. The algorithm-based strategy designs network parameter updating rule to alleviate performance degradation on previous tasks, such as GEM \cite{lopez2017gradient}, A-GEM \cite{chaudhry2018efficient} and OWM \cite{zeng2019continual}. 

\begin{figure}[!t]
	\centering
	\includegraphics[scale=0.8]{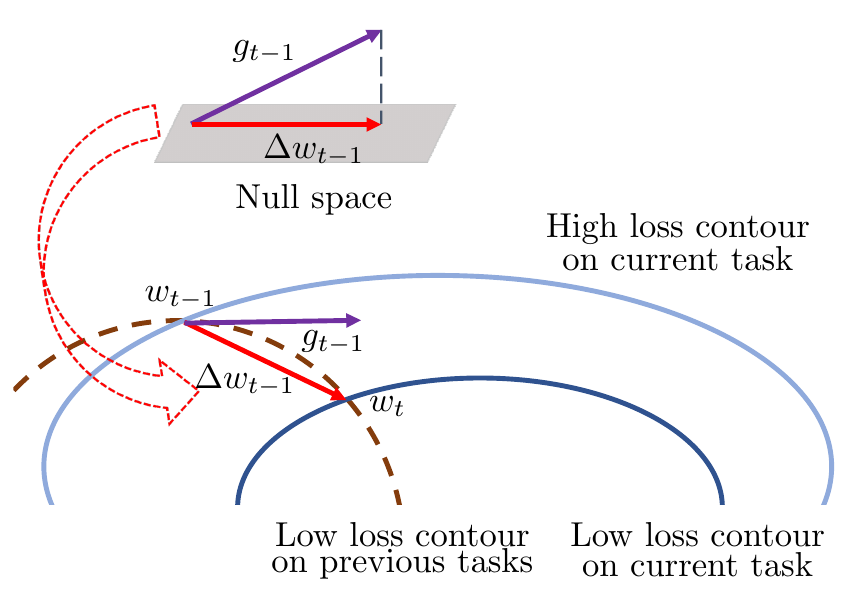}
	\caption{To avoid forgetting, we train network in the layer-wise null space of the corresponding uncentered covariance of all input features of previous tasks.}
	\label{fig:idea}
\end{figure}

In this paper, we focus on the setting of continual learning where the datasets from previous tasks are inaccessible. We first propose two theoretical conditions respectively for stability and plasticity of deep networks in continual learning. Based on them, we design a novel network training algorithm called Adam-NSCL for continual learning, which forces the network parameter update to lie in the null space of the input features of previous tasks at each network layer, as shown in \figref{fig:idea}. The layer-wise null space of input features can be modeled as the null space of the uncentered covariance of these features, which can be incrementally computed after learning each task. Since it is too strict to guarantee the existence of null space, we approximate the null space of each layer by the subspace spanned by singular vectors corresponding to smallest singular values of the uncentered covariance of input features. We embed this strategy into the Adam optimization algorithm by projecting the candidate parameter update generated by Adam \cite{kingma2015adam} into the approximate null space layer by layer, which is flexible and easy to implement. %
%along with the mini-batch network optimization process

We conduct various experiments on continual learning benchmarks in the setting that the datasets of previous tasks are unavailable, and results show that our Adam-NSCL is effective and outperforms the state-of-the-art continual learning methods. We also empirically verify the rationality of the approximate null space. 

The paper is organized as follows. We first introduce related works in \secref{sec:related}. In \secref{sec:method}, we present the mathematical conditions and then propose network training algorithm for continual learning in \secref{sec:appro}. In \secref{sec:exp}, we conduct experiments to verify the efficacy of our approach. 

\section{Related Work}\label{sec:related}
We next review the related works of continual learning in the following five categories. 

\textbf{Regularization-based strategy.} The basic idea of this strategy is to penalize the changes of network parameters when learning current task to prevent catastrophic forgetting. The typical methods include EWC \cite{kirkpatrick2017overcoming}, SI \cite{zenke2017continual}, MAS \cite{aljundi2018memory}, RWalk \cite{chaudhry2018riemannian} and NPC \cite{paik2020overcoming}. They impose regularization on the network parameters, and each network parameter is associated with an importance weight computed by different methods. These importance weights are required to be stored  for tasks in continual learning. Under the Bayesian framework, VCL \cite{nguyen2018variational}, CLAW \cite{adel2019continual} and IMM \cite{lee2017overcoming} take the posterior distribution of network parameters learned from previous tasks as the prior distribution of network parameters on current task, which implicitly penalizes the changes of network parameters under the Bayesian framework.

\textbf{Distillation-based strategy.} Inspired by knowledge distillation \cite{hinton2015distilling}, this strategy takes the network learned from previous tasks as teacher and the network being trained on current task as student, and then utilize a distillation term to alleviate performance degradation on previous tasks, such as LwF \cite{li2017learning}, GD-WILD \cite{lee2019overcoming}, lifelong GAN \cite{zhai2019lifelong}, MCIL \cite{liu2020mnemonics}, LwM \cite{dhar2019learning}, etc. Due to inaccessibility to the full datasets of previous tasks, they commonly use data of current task \cite{dhar2019learning,li2017learning}, external data \cite{lee2019overcoming}, coreset of previous tasks \cite{liu2020mnemonics,rebuffi2017icarl} or synthetic data \cite{zhai2019lifelong}, resulting in distributional shift \cite{2020Generative} to the original datasets.

\textbf{Replay-based strategy.} The replay-based strategy trains networks using both data of the current task and ``replayed'' data of previous tasks. Some existing works focus on selecting a subset of data from previous tasks~\cite{aljundi2019gradient,prabhu2020gdumb}, resulting in imbalance between the scale of datasets from current and previous tasks~\cite{wu2019large,zhao2020maintaining}. An alternative approach is to learn generative model to generate synthetic data to substitute the original data~\cite{hu2018overcoming,kemker2018fearnet,osta2019learning,shin2017continual}. They do not need to store data of previous tasks, however, the performance is significantly affected by the quality of generated data, especially for complex natural images.

\textbf{Architecture-based strategy.} In this strategy, the network architecture is dynamically modified by expansion or mask operation when encountering new tasks in continual learning. Methods of network expansion modify the network architecture by increasing network width or depth to break its representational limit when facing new tasks \cite{hung2019compacting,li2019learn,yoon2018lifelong}. This strategy may result in a powerful but redundant network that is computationally expensive and memory intensive. An alternative approach is to assign different sub-networks to different tasks by masking the neurons \cite{abati2020conditional,rajasegaran2019random,serra2018overcoming} or weights \cite{mallya2018piggyback}. The mask associated with each task needs to be learned and stored in the memory.

\textbf{Algorithm-based strategy.} This strategy performs continual learning from the perspective of network training algorithm. It focuses on designing network parameter updating rule to guarantee that network training on current task should not deteriorate performance on previous tasks. GEM \cite{lopez2017gradient} computes the parameter update by solving a quadratic optimization problem constraining the angle between the parameter update and the gradients of network parameters on data of previous tasks.  A-GEM ~\cite{chaudhry2018efficient} is an improved GEM  without solving quadratic optimization problem. A-GEM constrains that the network parameter update should be well aligned with a reference gradient computed from  a random batch of data from the previous tasks. Both of GEM and A-GEM need to store data of previous tasks. Different from GEM and A-GEM, OWM \cite{zeng2019continual} projects the parameter update into the orthogonal space of the space spanned by input features of each linear layer. The computation of the space projection matrix relies on the unstable inversion of matrix.

Our method called Adam-NSCL is a novel network training algorithm for continual learning, which trains networks in the approximate null space of feature covariance matrix of previous tasks to balance the network plasticity and stability. It does not require to design regularizers, revise network architecture and use replayed data in our method. Compared with OWM, Adam-NSCL relies on null space of feature covariance for achieving plasticity and stability with theoretical and empirical analysis, and overcomes the unstable matrix inversion in OWM. More discussions on the differences are in \secref{expres}.

\section{Analysis of  Stability and Plasticity}\label{sec:method}
In this section, we first present the preliminaries on the setting of continual learning, then propose mathematical conditions on the network parameter updates for the stability and plasticity, as the basis of our algorithm in \secref{sec:appro}. 
\subsection{Preliminaries}
In the setting of continual learning, a network $f$ with parameters $ \mathbf{w}$ is sequentially trained on a stream of tasks $\{\mathcal{T}_1, \mathcal{T}_2,\dots\}$, where task $\mathcal{T}_t$ is associated with paired dataset $\{X_t, Y_t\}$ of size $n_t$. The output of network $f$ on data $X_t$ is denoted as $f(X_t,\mathbf{w})$.

The initial parameters of network $f$ with $L$ linear layers on task $\mathcal{T}_t$ is set as $\tilde{\mathbf{w}}_{t-1}=\{\tilde{w}^1_{{t-1}}, \dots, \tilde{w}^L_{{t-1}}\}$ which is the optimal parameters after trained on task $\mathcal{T}_{t-1}$. When training $f$ on task $\mathcal{T}_t$ at the $s$-th training step, we denote the network parameters as $\mathbf{w}_{t,s}=\{w^1_{{t,s}},\dots, w^L_{{t,s}}\}$.  Correspondingly, the parameter update at the $s$-th training step on task $\mathcal{T}_t$ is denoted as $\Delta \mathbf{w}_{t,s}=\{\Delta w^1_{{t,s}},\dots, \Delta w^L_{{t,s}}\}$. When feeding data $X_p$ from task $\mathcal{T}_p$ $(p\leq t)$ to $f$ with optimal parameters $\tilde{\mathbf{w}}_{t}$ on task $\mathcal{T}_t$, the input feature and output feature at the $l$-th linear layer are denoted as $X^l_{{p,t}}$ and $O^l_{{p,t}}$, then  $$O^l_{{p,t}}=X^l_{{p,t}}  \tilde{w}^l_{{t}}, \ \ X^{l+1}_{{p,t}}=\sigma_l(O^l_{{p,t}})$$ with $\sigma_l$ as the nonlinear function and $X^1_{{p,t}}=X_p$. 

%For the convolutional layer, we can reformulate the output of convolution layer as the above matrix multiplication, which is unified with the fully-connected layer. Specifically, for each 3-D feature map, we flat each patch as a row vector, where the patch size is same as the corresponding 3-D convolutional kernel size, and the number of patches is the times that the kernel slides on the feature map when convolution. Then these row-wise vectors are concatenated to construct the 2-D feature matrix $X^l_{{p,t}}$. 3-D kernels at the same layer are flatted as column vectors of the 2-D parameter matrix ${w}^l_{{t,s}}$. 

For the convolutional layer, we can reformulate convolution as the above matrix multiplication, which is unified with the fully-connected layer. Specifically, for each 3-D feature map, we flat each patch as a row vector, where the patch size is same as the corresponding 3-D convolutional kernel size, and the number of patches is the times that the kernel slides on the feature map when convolution. Then these row-wise vectors are concatenated to construct the 2-D feature matrix $X^l_{{p,t}}$. 3-D kernels at the same layer are flatted as column vectors of the 2-D parameter matrix. 

\subsection{Conditions for continual learning}\label{sec:suff}

When being trained on the current task $\mathcal{T}_t$ without training data from previous tasks, the network $f$ is expected to perform well on the previous tasks, which is challenging since the network suffers from catastrophic forgetting. To alleviate the plasticity-stability dilemma, we propose two conditions for continual learning that guarantee the stability and plasticity respectively as follows. %As mentioned before, this is the \textit{plasticity-stability dilemma} \cite{mermillod2013stability}.

To derive Condition \ref{prop1} for network stability, we first present the condition of network parameter update to retain training performance on succeeding training tasks in Lemma \ref{lemma1} and Lemma \ref{lemma2}, depending on data of previous tasks. Then, we further propose the equivalent condition in Condition \ref{prop1}, free from storing data of previous tasks. After that, we present Condition \ref{prop2} for network plasticity.

\begin{lemma}\label{lemma1}
	Given the data $X_p$ from task $\mathcal{T}_p$,  and the network $f$ with $L$ linear layers is trained on task $\mathcal{T}_t$ ($t>p$). If network parameter update $\Delta w^l_{{t,s}}$ lies in the null space of $X^l_{{p,t-1}}$, \ie, 
	\begin{equation}\label{sc1}
		X^l_{p,t-1} \Delta w^l_{{t,s}} = 0,
	\end{equation}
	at each training step $s$, for the $l$-th layer of $f$ $(l=1,\dots,L)$,  we have $X^l_{{p,t}}=X^l_{p,t-1}$ and $f(X_p,\tilde{\mathbf{w}}_{t-1})=f(X_p,\tilde{\mathbf{w}}_{t})$.
\end{lemma}
\begin{proof}
	Please refer to the supplemental material. 
	%	 For every training step $t$, since $X^1_{p,c,t}=X^1_{p,c,t-1}$$=X_p$ holds, we have ${X}_{p,c,t}^1w^1_{c,t} = {X}_{p,c,t-1}^1 w^1_{c,t-1}$ if ${X}_{p,c-1,*}^1 \Delta w^1_{c,t} = 0$, where $w^1_{c,t} = w^1_{c,t-1} -\alpha \Delta w^1_{c,t}$ with $\alpha$ as learning rate. Furthermore, it is easy to verify that 
	%	 \begin{equation}
	%	 	{X}_{p,c-1,*}^1 w^1_{c-1,*} = {X}_{p,c,*}^1 w^1_{c,*},
	%	 \end{equation}
	% 	since $X^1_{p,c-1,*}=X^1_{p,c,*}=X_p$ and $w^1_{c-1,*}=w^1_{c,0}$. Therefore, 
	% 	\begin{equation}
	% 		{X}_{p,c,*}^2 = \sigma_1({X}_{p,c,*}^1 w^1_{c,*})=\sigma_1({X}_{p,c-1,*}^1 w^1_{c-1,*}) = {X}_{p,c-1,*}^2.
	% 	\end{equation}
	%	 Considering the recursive structure of network, we can prove that $X^l_{p,c,*}=X^l_{p,c-1,*}$ $(l=1,\dots,L)$ and $f(X_p,\mathbf{w}_{c,*})=f(X_p,\mathbf{w}_{c-1,*})$.
\end{proof}

Lemma \ref{lemma1} tells us that, when we train network on task $\mathcal{T}_t$, the network retains  its training loss on data $X_p$ in the training process, if the network parameter update satisfies Eqn.~(\ref{sc1}) at each training step. Considering that we initialize parameters $\mathbf{w}_{t,0}$ by $\tilde{\mathbf{w}}_{t-1}$, \ie, the optimal parameters on task $t-1$ for $t>1$, we have the following corollary.

\begin{cor}\label{cor1}
	Assume that network $f$ is sequentially trained on tasks $\{\mathcal{T}_1,\mathcal{T}_2, $ $ \cdots\}$. For each task $\mathcal{T}_t$ $(t>1)$ and $p<t$, if Eqn. \eqref{sc1} holds at every training step on task $\mathcal{T}_t$, we have $X^l_{{p,p}}=X^l_{p,t}$ $(l = 1, \cdots, L)$ and $f(X_p,\tilde{\mathbf{w}}_{t})=f(X_p,\tilde{\mathbf{w}}_{p})$.
\end{cor}

%In Corollary \ref{cor1}, Eqn. \eqref{sc1} should hold for any every task $\mathcal{T}_k$ $(k>1)$, therefore, after learning task $\mathcal{T}_{t-1}$, we have
%\begin{equation}\label{sc2}
%	X^l_{p,p}=X^l_{p,p+1}=\cdots=X^l_{p,t-1}
%\end{equation} 
%and $f(X_p,\tilde{\mathbf{w}}_{p})=f(X_p,\tilde{\mathbf{w}}_{p+1})=\cdots=f(X_p,\tilde{\mathbf{w}}_{t-1})$
%%\begin{equation}
%%	f(X_p,\mathbf{w}_{p})=f(X_p,\mathbf{w}_{p+1})=\cdots=f(X_p,\mathbf{w}_{c-1})
%%\end{equation}
%for any $p<t$. Furthermore, according to Lemma \ref{lemma1}, when training $f$ on task $\mathcal{T}_t$, if $\Delta w^l_{t,s}$ lies in the null space of $X^l_{p,t-1}=X^l_{p,p}$ at every layer $l$, we have $f(X_p,\tilde{\mathbf{w}}_{t})=f(X_p,\tilde{\mathbf{w}}_{t-1})=f(X_p,\tilde{\mathbf{w}}_{p})$, \ie, the performance of $f$ on task $\mathcal{T}_p$ does not become worse after learning task $\mathcal{T}_t$ for any $p<t$.

%In the context of continual learning, it is supposed that the performance of $f$ on task $\mathcal{T}_p$ does not become worse after learning task $\mathcal{T}_t$ for all $p<t$. Therefore, according to Eqns. \eqref{sc1} and \eqref{sc2}, $\Delta w^l_{t,s}$ should lie in the intersection of null spaces of $X^l_{{p,p}}$ $(p=1,\dots,t-1)$. By denoting $\bar{X}^l_{t-1}$$=[{X_{1,1}^l}^{\top},\dots,{X_{t-1,t-1}^l}^{\top}]^\top$, we have the following lemma.

Corollary \ref{cor1} suggests that the training loss on data $X_p$ is retained if the network trained on the following tasks satisfies Eqn.~(\ref{sc1}) and network parameters at each task are initialized by the trained network of the last task.

We further denote $ \bar{X}^l_{t-1}=[{X_{1,1}^l}^{\top},\cdots,{X_{t-1,t-1}^l}^{\top}]^\top$, which is the concatenation of input features of  $l$-th network layer on each task data $X_p$ ($p < t$) using trained network parameters on task $\mathcal{T}_p$. Then the following lemma holds.

\begin{lemma}\label{lemma2}
	Assume that $f$ is being trained on task $\mathcal{T}_t$ ($t>1$). If $\Delta w^l_{{t,s}}$ lies in the null space of $\bar{X}^l_{t-1}$ at each training step $s$, \ie,
	\begin{equation}
		\bar{X}^l_{t-1} \Delta w^l_{{t,s}}=0,
	\end{equation}
	for $l=1,\cdots,L$, we have $f(X_p,\tilde{\mathbf{w}}_{t})=f(X_p, \tilde{\mathbf{w}}_{p})$ for all $p=1,\cdots,t-1$.
\end{lemma}

Lemma \ref{lemma2} guarantees the stability of $f$. However, it is inefficient since it requires to store all features $X^l_{{p,p}}$ of $f$ for all $p<t$, which is memory-prohibited. To overcome this limitation, we propose the following Condition \ref{prop1} based on uncentered feature covariance $\bar{\mathcal{X}}^l_{t-1}\triangleq\frac{1}{\bar{n}_{t-1}} (\bar{X}^{l}_{t-1})^\top \bar{X}^l_{t-1}$  to guarantee stability, where $\bar{n}_{t-1}$ is the total number of seen data, \ie, the number of rows of $\bar{X}^l_{t-1}$.% based on the fact that the null space of $\bar{X}_{l_c}$ equals to null space of the uncentered covariance of $\bar{X}_{l_c}$.
\begin{prop}[stability]	\label{prop1}
	When $f$ is being trained on task $\mathcal{T}_t$,  $\Delta w^l_{{t,s}}$ at each training step $s$ should lie in the null space of the uncentered  feature covariance matrix $\bar{\mathcal{X}}^l_{t-1}$  for $l = 1, \cdots, L$, \ie, 
	\begin{equation}\label{eqcond1}
		\bar{\mathcal{X}}^l_{t-1} \Delta w^l_{{t,s}}=0.
	\end{equation}
\end{prop}
It is easy to verify that the null space of $\bar{X}^l_{t-1}$ equals to null space of the uncentered feature covariance of $\bar{\mathcal{X}}^l_{t-1}$. Therefore, if Condition \ref{prop1} holds, we have $f(X_p,\tilde{\mathbf{w}}_{t})=f(X_p,\tilde{\mathbf{w}}_{p})$ for all $p<t$ according to Lemma \ref{lemma2}, \ie, the performance of $f$ on previous task data will not be degraded after learning current task.

\textit{\textbf{Memory analysis.}} The memory consumption of storing $\bar{\mathcal{X}}^l_{t}$ is fixed, irrelevant to the number of tasks and data. Specifically, if we denote the dimension of the feature at layer $l$ as $h^l$, the size of $\bar{X}^l_{t}$ is $\bar{n}_{t} \times h^l$ (usually $\bar{n}_{t}\gg h^l$), while the size of $\bar{\mathcal{X}}^l_{t}$ is $h^l \times h^l$. Therefore, Condition \ref{prop1} supplies a more memory efficient way to guarantee the stability of $f$ than Lemma \ref{lemma2}. 

Condition \ref{prop1} guarantees the stability of network in continual learning. The other requirement of continual learning is the plasticity of $f$ concerning the ability to obtain new knowledge from current task. Condition \ref{prop2} will  provide the condition that guarantees the plasticity of network $f$.
\begin{prop}[plasticity]
	\label{prop2}
	Assume that the network $f$ is being trained on task $\mathcal{T}_t$, and $\mathbf{g}_{t,s}=\{g^1_{t,s},\dots,g^L_{t,s}\}$ denotes the parameter update generated by a gradient-descent training algorithm for training $f$ at training step $s$. $\langle\Delta \mathbf{w}_{t,s}, \mathbf{g}_{t,s}\rangle > 0$  should hold where $\langle \cdot, \cdot\rangle$ represents inner product.
\end{prop}
If parameter update  $\Delta\mathbf{w}_{t,s}$ satisfies Condition \ref{prop2} when training $f$ on task $\mathcal{T}_t$, the training loss after updating parameters using $\Delta\mathbf{w}_{t,s}$ will decrease, \ie, the network can be trained on this task. Please see supp. for proof.

\begin{figure}[!tp]
	\centering
	\includegraphics[scale=0.75]{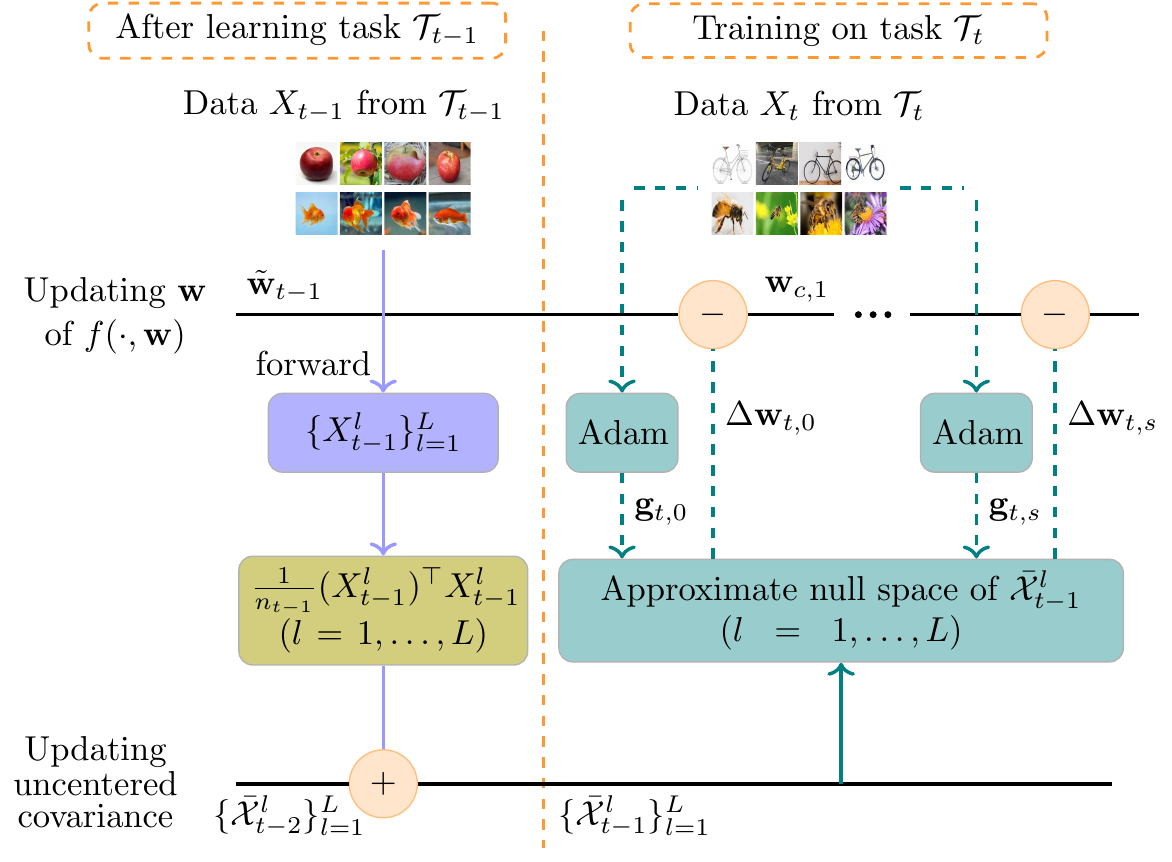}
	\caption{The pipeline of our algorithm.}
	\label{figalg}
\end{figure}
\section{Network Training in Covariance Null Space}\label{sec:appro}
In this section, we propose a novel network training algorithm called Adam-NSCL for continual learning based on Adam and the proposed two conditions respectively for stability and plasticity. In continual learning, we maintain layer-wise feature covariance matrices, which are incrementally updated using features of the coming task. Given the current task, starting from the network trained on previous tasks, we update the network parameters on current task using Adam-NSCL where the candidate parameter update generated by Adam is projected into the approximate null space of corresponding feature covariance matrix layer by layer to balance the network stability and plasticity. 
%In this section, with the proposed two conditions respectively for stability and plasticity, we propose a novel network training algorithm called Adam-NSCL for continual learning based on Adam optimization algorithm, a variant of stochastic gradient descent. In the continual learning, we maintain layer-wise feature covariance matrices, iteratively updated using features on previous tasks. Given a new task, starting from the network trained on previous tasks, we update the network parameters on this new task using a revised Adam algorithm where the candidate update generated by Adam is projected into the approximate null space of corresponding feature covariance matrix layer by layer to balance the network stability and plasticity. 

Figure \ref{figalg} illustrates the pipeline of our proposed continual learning algorithm. After learning task $\mathcal{T}_{t-1}$, we feed $X_{t-1}$ to the learned network $f(\cdot,\tilde{\mathbf{w}}_{t-1})$ to get the input feature $X^l_{t-1}$ at each layer. Then we compute the uncentered covariance ${\mathcal{X}}^l_{t-1}=\frac{1}{n_{t-1}} (X^l_{t-1})^\top X^l_{t-1}$  with $n_{t-1}$ as the number of data on task $\mathcal{T}_{t-1}$. Subsequently, we update the uncentered feature covariance matrix for each layer by
\begin{equation}
	\bar{\mathcal{X}}^l_{t-1}=\frac{\bar{n}_{t-2}}{\bar{n}_{t-1}} \bar{\mathcal{X}}^l_{t-2}+ \frac{n_{t-1}}{\bar{n}_{t-1}} {\mathcal{X}}^l_{t-1},
\end{equation}
with $\bar{n}_{t-1}=\bar{n}_{t-2} + n_{t-1}$ as the total number of seen data.
Following that, we compute the approximate null space of $\bar{\mathcal{X}}^l_{t-1}$. When training network with $\tilde{\mathbf{w}}_{t-1}$ as initialization on task $\mathcal{T}_t$, we first utilize Adam to generate a candidate parameter update $\mathbf{g}_{t,s}$ at $s$-th step ($s=1,2,\dots$), then project $\mathbf{g}_{t,s}$ into the approximate null space of $\bar{\mathcal{X}}^l_{t-1}$ layer by layer to get the parameter update $\Delta \mathbf{w}_{t,s}$. 

In the following, we first introduce derivation of the approximate null space in \secref{sec:appns}, and discuss the projection satisfying Conditions \ref{prop1} and \ref{prop2}. Subsequently, we present our proposed continual learning algorithm in \AlgRef{alg} and the way to find the approximate null space in \AlgRef{alg:null}.

\subsection{Approximate null space}\label{sec:appns}
According to Condition \ref{prop1}, for the stability of network in continual learning, we can force the parameter update to lie in the null space of uncentered covariance of all previous input features at each network layer. However, it is too strict to guarantee the existence of null space. Therefore, we propose \AlgRef{alg:null} to find the approximate null space based on SVD of the uncentered covariance matrix. 

By applying SVD to $\bar{\mathcal{X}}^l_{t-1}$, we have 
\begin{equation}
	U^l, \Lambda^l, {(U^{l})^\top}=\text{SVD}(\bar{\mathcal{X}}^l_{t-1}),
\end{equation}
where $U^l=[U^l_{1}, U^l_{2}]$ and $\Lambda^l=\begin{bmatrix}
	\Lambda^l_{1} & 0\\
	0&\Lambda^l_{2}
\end{bmatrix}$. If all singular values of zero are in $\Lambda^l_{2}$, \ie, $\Lambda^l_{2}=0$, then $\bar{\mathcal{X}}^l_{t-1} U^l_{2} = U^l_{1}\Lambda^l_{1} (U^l_{1})^\top U^l_{2}=0$ holds, since $U^l$ is an unitary matrix. It suggests that the range space of $U^l_{2}$ is the null space of $\bar{\mathcal{X}}^l_{t-1}$. Thus we can get the parameter update $\Delta w^l_{{t,s}}$ lying in the null space of $\bar{\mathcal{X}}^l_{t-1}$ by
\begin{equation} \label{update}
	\Delta w^l_{{t,s}}=U^l_{2} (U^l_{2})^\top g^l_{{t,s}}
\end{equation}
with $U^l_{2} (U^l_{2})^\top$ as projection operator \cite[Eqn. (5.13.4)]{meyer2000matrix}. Thus we get $\Delta w^l_{{t,s}}$ satisfying Condition \ref{prop1}.
\begin{figure}[!bp]
	\centering
	\includegraphics[scale=0.5]{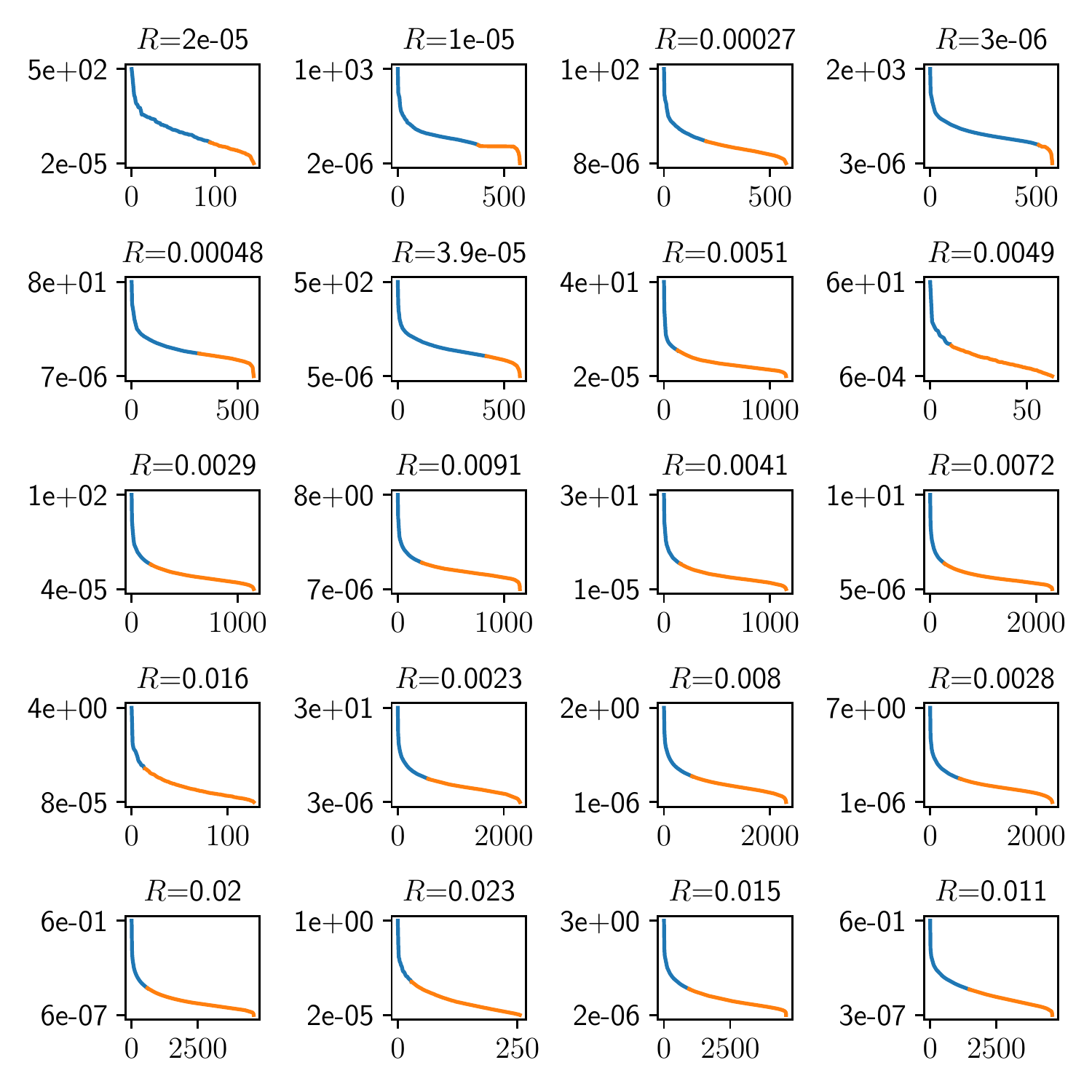}
	\caption{Singular values of uncentered covariance matrix at different layers of pretrained ResNet-18 on ImageNet ILSVRC 2012. Orange curves denote the singular values smaller than $50\lambda^l_{\text{min}}$.}
	\label{svf}
\end{figure}
\begin{algorithm}[!t]
	\caption{Adam-NSCL for continual learning} 
	\label{alg}
	%	\textbf{Inputs:} Datasets $\{{X}_t, {Y}_t\}$ for task $\mathcal{T}_t\in \{\mathcal{T}_1, \mathcal{T}_2,\dots\}$; network $f(\cdot, \mathbf{w})$ with $L$ linear layers; learning rate $\alpha$.  \\
	\textbf{Inputs:} Datasets $\{{X}_t, {Y}_t\}$ for task $\mathcal{T}_t\in \{\mathcal{T}_1, \mathcal{T}_2,\dots\}$; network $f(\cdot, \mathbf{w})$ with $L$ linear layers; learning rate $\alpha$.  \\
	\textbf{Initialization:} Initialize $\tilde{\mathbf{w}}_{0}$ randomly, $\bar{\mathcal{X}}_0^l=0$, number of seen data $\bar{n}_0=0$.
	\begin{algorithmic}[1]
		\STATE{\textit{\# sequential tasks}}
		\FOR{task $\mathcal{T}_t\in \{\mathcal{T}_1, \mathcal{T}_2,\dots\}$}
		\IF{$t>1$}
		\STATE{\textit{\# compute the approximate null space}}
		\STATE{Get $U^l_{2}$, $\bar{\mathcal{X}}^l_{{t-1}}$ and $\bar{n}_{t-1}$  $(l=1,\dots,L)$ by \AlgRef{alg:null} with $\{{X}_{t-1}, {Y}_{t-1}\}$,  $f(\cdot, \tilde{\mathbf{w}}_{t-1})$, $\bar{\mathcal{X}}^l_{{t-2}}$ and $\bar{n}_{t-2}$ as inputs.}
		\ENDIF
		\STATE{\textit{\# train $f(\cdot, \mathbf{w})$ on task $\mathcal{T}_t$.}}
		\STATE{Set $s=0$ and $\mathbf{w}_{t,0}=\tilde{\mathbf{w}}_{t-1}$;}
		\WHILE{not converged}
		\STATE{Sample a batch $\{\mathbf{x}, \mathbf{y}\}$ from $\{{X}_t, {Y}_t\}$.}
		\STATE{Compute $f(\mathbf{x}, \mathbf{w}_{t,s})$, then get candidate parameter update $\mathbf{g}_{t,s}=\{g^1_{{t,s}},\dots,g^L_{{t,s}}\}$ by Adam.}
		%		\STATE{Compute $g^l_{c,t}$ by Adam.}
		\IF{$t=1$}
		\STATE{$\Delta w^l_{{t,s}}=g^l_{{t,s}}$, $l=1,\dots, L$}
		\ELSE
		%		\STATE{\textit{\# projecting $g^l_{{t,s}}$ into the approximate null space}}
		\STATE{$\Delta w^l_{{t,s}}=U^l_{2} (U^l_{2})^\top g^l_{{t,s}}$, $l=1,\dots, L$}
		\ENDIF
		%		\STATE{$\Delta w^l_{c,t}=g^l_{c,t}$ if $c=1$ else $U_2^lU_2^{l'}g^l_{c,t}$.}
		\STATE{$w^l_{{t,s+1}}=w^l_{{t,s}} - \alpha \Delta w^l_{{t,s}}$, $l=1,\dots, L$}
		\STATE{$s=s+1$}
		\ENDWHILE
		%		\STATE{Denote the optimal parameter on $\mathcal{T}_c$ as $\mathbf{w}_{c}$ .}
		\ENDFOR
	\end{algorithmic}
\end{algorithm}

However, it is unrealistic to guarantee that there exists zero singular values. Inspired by Principal Component Analysis, if considering $U^l_{1}$ as principal components, $\bar{\mathcal{X}}^l_{t-1}$ can be approximated by $U^l_{1}\Lambda^l_{1} (U^l_{1})^\top$, which indicates that $\bar{\mathcal{X}}^l_{t-1}  U^l_{2} \approx U^l_{1}\Lambda^l_{1} (U^l_{1})^\top U^l_{2}=0$,
%\begin{equation}
%\mathcal{X}^l_{t}  U^l_{2} \approx U^l_{1}\Lambda^l_{1} (U^l_{1})^\top U^l_{2}=0,
%\end{equation}
\ie, we can take the range space of $U^l_{2}$ as the approximate null space of $\bar{\mathcal{X}}^l_{t-1}$ where $U^l_{2}$ corresponds to the smallest singular values in $\Lambda^l_{2}$. We adaptively select $\Lambda^l_{2}$ with diagonal singular values $\lambda \in \{\lambda|\lambda \leq a \lambda^l_{{\text{min}}}\}$ $(a>0)$, where $\lambda^l_{{\text{min}}}$ is the smallest singular value. Furthermore, to empirically verify the rationality of the approximation, we utilize the proportion $R$ of $\bar{\mathcal{X}}^l_{t-1}$ explained by $U^l_{2}$ \cite{jolliffe2016principal} as
\begin{equation}\label{ratio}
	R=\frac{\Sigma_{\lambda\in \mbox{diag}\{\Lambda^l_{2}\}}\lambda}{\Sigma_{\lambda\in \mbox{diag}\{\Lambda^l\}}\lambda},
\end{equation}
where ``$\mbox{diag}$'' denotes the diagonal elements. If $R$ is small, the sum of singular values of $\Lambda^l_2$ is negligible, suggesting that it is reasonable to approximate null space of uncentered covariance matrix by the range space of $U^l_2$.

%	$\{\bar{\mathcal{X}}^l_{{t-1}}\}_{l=1}^L$, $\bar{n}_{t-1}$.
\begin{algorithm}[!t]
	\caption{Updating the uncentered covariance incrementally and computing the null space.} 
	\label{alg:null}
	\textbf{Inputs:} Dateset $\{{X}_{t-1}, {Y}_{t-1}\}$ of size $n_{t-1}$ for task $\mathcal{T}_{t-1}$; network $f(\cdot, \tilde{\mathbf{w}}_{t-1})$; $\{\bar{\mathcal{X}}^l_{{t-2}}\}_{l=1}^L$; hyperparameter $a>0$; number of seen data $\bar{n}_{t-2}$.\\
	\textbf{Output:} $U^l_2$; $\{\bar{\mathcal{X}}^l_{{t-1}}\}_{l=1}^L$; $\bar{n}_{t-1}$
	\begin{algorithmic}[1]
		\STATE{\textit{\# Compute the uncentered covariance on task $\mathcal{T}_{t-1}$}}
		\STATE{Initialize uncentered covariance matrices $\mathcal{X}^l_{t-1}$ on task $\mathcal{T}_{t-1}$ as 0 for $l=1,\dots,L$.}
		\FOR{batch $\{\mathbf{x}, \mathbf{y}\}$ from $\{{X}_{t-1}, {Y}_{t-1}\}$}
		\STATE{Get the input feature $\mathbf{x}^l$ at the $l$-th layer $(l=1,\dots,L)$ by forward propagating $\mathbf{x}$ on $f(\cdot,\tilde{\mathbf{w}}_{t-1})$.\\}
%		\FOR {$l=1,\dots,L$}
%		\STATE{${\mathcal{X}}^l_{t-1} ={\mathcal{X}}^l_{t-1} + (\mathbf{x}^l)^\top\mathbf{x}^l$.}				
%		\ENDFOR
		\STATE{${\mathcal{X}}^l_{t-1} ={\mathcal{X}}^l_{t-1} + (\mathbf{x}^l)^\top\mathbf{x}^l$ for $l=1,\dots,L$.}			
		\ENDFOR
		\STATE{${\mathcal{X}}^l_{t-1} = \frac{1}{n_{t-1}} {\mathcal{X}}^l_{t-1}$ for $l=1,\dots, L$.}
		\STATE{$\bar{n}_{t-1}=\bar{n}_{t-2}+n_{t-1}$.}
%		\FOR {$l=1,\dots,L$}
		%			\STATE{Computing the null space of $\mathcal{X}_c^l$ as introduced in \secref{sec:sol}.}	
		%		\STATE{Obtain $U_2^l$ by applying SVD to $\mathcal{X}_c^l$ as introduced in \secref{sec:sol}.}
		\STATE{\textit{\# Update the uncentered covariance $\bar{\mathcal{X}}^l_{t-2}$.}}
		\STATE{$\bar{\mathcal{X}}^l_{t-1}=\frac{\bar{n}_{t-2}}{\bar{n}_{t-1}} \bar{\mathcal{X}}^l_{t-2}+ \frac{n_{t-1}}{\bar{n}_{t-1}} {\mathcal{X}}^l_{t-1}$, $l=1,\dots,L$.}
		\STATE{\textit{\# Compute the approximate null space for each layer}}
		\STATE{$U^l, \Lambda^l, {(U^{l})^\top}=$SVD($\bar{\mathcal{X}}^l_{t-1}$).}
		\STATE{Get $\Lambda^l_{2}$ with diagonal singular values $\lambda \in \{\lambda|\lambda<a\lambda^l_{\text{min}}\}$ where $\lambda^l_{\text{min}}$ is the smallest singular value.}	
		\STATE{Get singular vectors $U^l_{2}$ that correspond to $\Lambda^l_{2}$.}
%		\ENDFOR
	\end{algorithmic}
\end{algorithm}

\textbf{Example}. We take the pretrained ResNet-18 on dataset of ImageNet ILSVRC 2012 \cite{ILSVRC15} as example. Figure \ref{svf} shows the curves of singular values of uncentered covariance matrix $\bar{\mathcal{X}}^l_{t}$ of each linear layer indexed by $l$ with $a=50$. All proportions $R$ of different layers are smaller than 0.05, indicating that the selected $U^l_{2}$ corresponding to smallest singular values is negligible to explain $\bar{\mathcal{X}}^l_{t}$. Therefore, it is reasonable to approximate the null space by the range space of $U^l_{2}$.%Additionally, we report $R$ with different setting of $a$ in experiments, where almost all $R$ are smaller than 0.05. Therefore, it is rational to take the range space of $U^l_{2}$ as the approximate null space.

As a summary, for a novel task $\mathcal{T}_t$, our continual learning algorithm (shown in \figref{figalg}) projects parameter update $\mathbf{g}_{t, s}=\{g^l_{t,s}\}_{l=1}^L$ at $s$-th training step generated by Adam into the approximate null space layer by layer, and get the parameter update $\Delta \mathbf{w}_{t,s}=\{\Delta {w}^l_{t,s}\}_{l=1}^L$ following Eqn.~(\ref{update}).  We prove that $\langle\Delta \mathbf{w}_{t,s}, \mathbf{g}_{t,s}\rangle  \geq0$ always holds, which can be found in the supplemental material. To guarantee that the network can be trained on task $\mathcal{T}_t$ using the above parameter updating rule, it is supposed that $\langle\Delta \mathbf{w}_{t,s}, \mathbf{g}_{t,s}\rangle  >0$ as discussed in Condition \ref{prop2}. We empirically found that it holds in our experiments and our algorithm can succeed in decreasing training losses on sequential training tasks.

Our Adam-NSCL is summarized in \AlgRef{alg}. The training process is to loop over incoming tasks, and task $\mathcal{T}_t$ ($t>1$) is trained by Adam with gradients projected into the null space of accumulated covariance (line 15 in \AlgRef{alg}). The null space is obtained by \AlgRef{alg:null} after learning task $\mathcal{T}_{t-1}$. In  \AlgRef{alg:null}, we first feed all training data of task $\mathcal{T}_{t-1}$ to accumulate covariance in lines 3-8, and update the uncentered covariance in line 10. Then we compute the approximate null space in lines 12-14 by SVD.

The hyperparameter $a$ controls the balance of stability and plasticity. Larger $a$ suggests that we use larger approximate null space $U_2^l$ covering more small singular values, then the null space assumption in Eqn. (\ref{eqcond1}) hardly holds, reducing the stability in continual learning. On the other hand, larger $a$ leads to larger approximate null space, enabling us to update network parameters in a larger null space, increasing the plasticity to learn knowledge on current task.

%\subsection{Algorithm}\label{sec:sol}
%Based on \secref{sec:appro}, we design our algorithm as in \AlgRef{alg}, the pipeline of which is shown in \figref{figalg}. When training network $f(\cdot, \mathbf{w})$ on task $\mathcal{T}_t$ $(t>1)$, for the parameter at the $l$-th layer $(l=1,\dots,L)$, we first generate a candidate update $\mathbf{g}_{t,s}=\{g^1_{{t,s}},\dots,g^L_{{t,s}}\}$  by Adam, then get the corresponding parameter update $\Delta \mathbf{w}_{t,s}=\{\Delta w^1_{{t,s}},\dots, \Delta w^L_{{t,s}}\}$ by projecting $g^l_{{t,s}}$ into the approximate null space specified by $U^l_{2}$ with projection operator $U^l_{2} (U^l_{2})^\top$ \cite[Eqn. (5.13.4)]{meyer2000matrix} for every layer $l$, where $U^l_{2}$ can be obtained by applying SVD to the uncentered covariance $\mathcal{X}^l_{t-1}$ of all previous input features at the $l$-th layer as described in \AlgRef{alg:null}. The uncentered covariance $\mathcal{X}^l_{t}$  is incrementally update after learning from task $\mathcal{T}_t$ as in \AlgRef{alg:null},
%\begin{equation}
%	\mathcal{X}^l_{t}=\mathcal{X}^l_{t-1}+  \bar{\mathcal{X}}^l_{t}
%\end{equation}
%with $\bar{\mathcal{X}}^l_{t} = \frac{1}{n^l} (X^l_{t})^\top X^l_{t}$,
%where $X^l_{t}$ is the input feature at $l$-th linear layer of $f(X_{t}, \tilde{\mathbf{w}}_t)$, and $n^l$ is the number of rows of $X^{l}_t$. $\bar{\mathcal{X}}^l_{t}$  can be incrementally computed batch-wise.

\section{Experiments}\label{sec:exp}
We apply Adam-NSCL algorithm to different sequential tasks for continual learning\footnote{https://github.com/ShipengWang/Adam-NSCL}. After introducing experimental setting, we show the results compared with SOTA methods, following which we empirically analyze our algorithm.
\subsection{Experimental setting}\label{sec:expset}
We first describe the experimental settings on datasets, implementation details, metrics and compared methods.
%\subsubsection{Datasets}

\textbf{Datasets}. We evaluate on continual learning datasets, including 10-split-CIFAR-100, 20-split-CIFAR-100 and 25-split-TinyImageNet. {10-split-CIFAR-100} and {20-split-CIFAR-100} are constructed by splitting CIFAR-100 \cite{krizhevsky2009learning}  into 10 and 20 tasks, and the classes in different tasks are disjoint. {25-split-TinyImageNet} is constructed by splitting TinyImageNet \cite{wu2017tiny} containing $64\times64$ RGB images into 25 tasks, which is a harder setting due to longer sequence.
%\begin{itemize}
%	\item \textbf{10-split-CIFAR100}: CIFAR-100 is split into 10 tasks, where each task has 10 disjoint classes.
%	\item \textbf{20-split-CIFAR100}: It is similar to 10-split-CIFAR-100 but where CIFAR-100 is split into 20 tasks.
%	\item \textbf{25-split-TinyImageNet}: TinyImageNet, consisting of $64\times64$ RGB images, is split into 25 tasks with 8 classes per task, where the classes are disjoint in different tasks.
%\end{itemize}
%\subsubsection{Implementation details of our method}

\textbf{Implementation details}. Adam-NSCL is implemented using PyTorch \cite{pytorch}. We take ResNet-18 \cite{he2016identity} as the backbone network in our experiments. All tasks share the same backbone network but each task has its own classifier. The classifier will be fixed after the model is trained on the corresponding task. For batch normalization layer, we regularize its parameters using EWC \cite{kirkpatrick2017overcoming}. The learning rate starts from $5\e{-5}$ and decays at epoch 30, 60 with a multiplier 0.5 (80 epochs in total).  The batch size is set to 32 for 10-split-CIFAR-100 and 16 for the other two datasets. The regularizer coefficient of EWC for penalizing parameters at batch normalization is set to 100. At each linear layer, to approximate null space of uncentered covariance, we set $a=30$ for 20-split-CIFAR-100 while $a=10$ for the other two datasets. We also study the effect of $a$ in \secref{sec:ana}.

\textbf{Compared methods}. We compare our method with various continual learning methods including \textit{EWC} \cite{kirkpatrick2017overcoming}, \textit{MAS} \cite{aljundi2018memory}, \textit{MUC-MAS} \cite{muc2020liu}, \textit{SI} \cite{osta2019learning},  \textit{LwF} \cite{li2017learning}, \textit{GD-WILD} \cite{lee2019overcoming}, \textit{GEM},  \cite{lopez2017gradient}, \textit{A-GEM} \cite{chaudhry2018efficient}, \textit{MEGA} \cite{guo2020improved}, InstAParam \cite{mitigating} and \textit{OWM} \cite{zeng2019continual}. For a fair comparison, the backbone networks employed in these methods are all ResNet-18. \textit{EWC}, \textit{MAS}, \textit{MUC-MAS} and \textit{SI} regularize the changes of parameters across tasks, where each parameter is associated with a weight of importance. \textit{LwF} and \textit{GD-WILD} are based on knowledge distillation using different dataset for distillation to preserve learned knowledge on previous tasks. GEM, A-GEM, \textit{MEGA} and OWM focus on designing network training algorithm to overcome forgetting. InstAParam is based on architecture-based strategy. Among these methods, EWC, MAS, MUC-MAS and SI need to store the importance weight in memory, GD-WILD, GEM A-GEM and MEGA need to store data of previous tasks, and OWM needs to store the projection matrix which is incrementally computed with an approximate inversion of matrix.

%We now discuss the difference between our algorithm and OWM\cite{zeng2019continual}. Though both of OWM and ours are based on projecting gradient into the null space, the way to construct the null space is not the same. The null space of OWM is specified by the features directly and the projection matrix is induced by the features. They update the projection matrix incrementally with an approximate inversion of matrix, where the approximate error would be accumulated. However, the null space of ours is specified by the feature covariance and the the projection matrix is computed by SVD. We incrementally update the covariance with no approximate error. Therefore, our algorithm outperforms OWM as the experimental results show.

%All of the above methods need to save the network learned on old tasks when learning from new task, which is not necessary for the algorithm-based strategy including GEM \cite{lopez2017gradient}, A-GEM \cite{chaudhry2018efficient}, OWM \cite{zeng2019continual} and our method. However, GEM \cite{lopez2017gradient} and A-GEM \cite{chaudhry2018efficient} demand a memory buffer to store the previous data, which is not in keeping with the setting of continual learning. Additionally, GEM needs to solve a Quadratic Problem at every training step. Though OWM \cite{zeng2019continual} does not have to store previous data, it requires to compute the unstable inversion of matrix, where the error is accumulated. OWM also does not work for batch normalization layer, thus EWC is utilized to penalize the variation of parameters at batch normalization layer.

\textbf{Evaluation metrics}. We employ the evaluation metrics proposed in \cite{lopez2017gradient}, including backward transfer (BWT) and average accuracy (ACC). BWT is the average drop of accuracy of the network for test on previous tasks after learning current task. Negative value of BWT indicates the degree of forgetting in continual learning. ACC is the average accuracy of the network on test datasets of all seen tasks. With similar ACC, the one having larger BWT is better.

%Additionally, we compare the burden of memory buffer to store necessary task information. Our approach needs to store the uncentered covariance for every layer, OWM needs to store the projection matrix, while EWC, MAS, SI and MUC-MAS need to store the weight of importance for each parameter. For fairly comparison, we report the \textit{Memory Ratio} which is defined as the ratio of memory buffer and model size.
%\begin{equation}
%	\text{Memory Ratio} = \frac{\text{the size of memory buffer}}{\text{the size of the model}}
%\end{equation}

\subsection{Experimental results}\label{expres}
We next compare different continual learning algorithms. The details on the comparative results on three datasets are presented as follows.
\begin{table}[!tbp]
	\begin{center}
		\scalebox{0.89}{
			\begin{tabular}{l|cc}
				\toprule
				\multirow{2}{5pt}{Method}
				&  \multicolumn{2}{|c}{10-split-CIFAR-100}   \\
				\cmidrule{2-3}
				&ACC (\%) & BWT(\%)  \\
				\midrule
				EWC \cite{kirkpatrick2017overcoming}   &70.77&  -2.83  \\
				MAS  \cite{aljundi2018memory}  &66.93&  -4.03   \\
				MUC-MAS \cite{muc2020liu}  &63.73&  -3.38\\
				SI \cite{osta2019learning}   &60.57&  -5.17\\
				LwF \cite{li2017learning}  &70.70&  -6.27 \\
				InstAParam \cite{mitigating} & 47.84& -11.92\\
				${}^*$GD-WILD \cite{lee2019overcoming}  & 71.27&  -18.24  \\%&$-$\\
				${}^*$GEM \cite{lopez2017gradient}  &49.48&  {2.77} \\
				${}^*$A-GEM \cite{chaudhry2018efficient} &49.57&  -1.13  \\%&$-$\\
				${}^*$ MEGA \cite{guo2020improved} &54.17&-2.19\\
				OWM	\cite{zeng2019continual} &68.89&  -1.88\\%&823\\
				\midrule
				Adam-NSCL   &\textbf{73.77}&  -1.6 \\%&823\\ %58.82 -8.17
				\bottomrule
			\end{tabular}	
		}
	\end{center}
	\caption{Comparisons of ACC and BWT for ResNet-18 sequentially trained on 10-split-CIFAR-100 using different methods. Methods required to store previous data are denoted by ${}^*$.} % Methods required to store previous data are denoted by ${}^*$.
	\label{final10}
\end{table}

%\begin{table}[!htbp]
%	\begin{center}
%		\scalebox{0.8}{
%			\begin{tabular}{l|ccccc}
%				\hline\hline
%				$l$&1     & 2      & 3     & 4       & 5   \\
%				%				cond &  4.4\e{5}    & 8.0\e{6}  &  5.8\e{5} & 1.0\e{6} & 3.1\e{5}\\
%				$R$   &  5.2\e{-5}  & 5.9\e{-5}  & 1.6\e{-3}    & 9\e{-4}&3.7\e{-3}\\
%				\hline\hline
%				$l$& 6      &7 &8     & 9      & 10 \\
%				%				cond & 4.3\e{5}  & 2.1\e{5}  & 7.5\e{3}  &  2.1\e{5}  &2.4\e{5}\\
%				$R$   & 2.6\e{-3}   &1.2\e{-2}   & 1.8\e{-2}  & 1.2\e{-2}  & 1.2\e{-2}  \\
%				\hline\hline
%				$l$   & 11       & 12   & 13      &14 &15  \\
%				%				cond  & 2.3\e{5}  & 4.4\e{5}  & 1.0\e{4}  &  4.3\e{5} &   5.9\e{5}\\
%				$R$   & 1.1\e{-2}   &1.2\e{-2}   &2.5\e{-2}   &1.2\e{-2}   &8.3\e{-3}    \\
%				\hline\hline
%				$l$   & 16      & 17     & 18       & 19   & 20       \\
%				%				cond  &  7.0\e{5} & 3.6\e{6}  &3.9\e{4}  &  3.5\e{6} &3.2\e{7}   \\
%				$R$   &   7.1\e{-3}      & 1.8\e{-3}  &1.4\e{-2}   &2.0\e{-3}   & 6.3\e{-5}   \\
%				\hline\hline
%			\end{tabular}
%		}
%	\end{center}
%	\caption{The ratio $R$ at each layer for network trained on 10-split-CIFAR-100.}
%	\label{exp1cond}
%\end{table}

\begin{figure}[!b]
	\centering
	\includegraphics[scale=0.45]{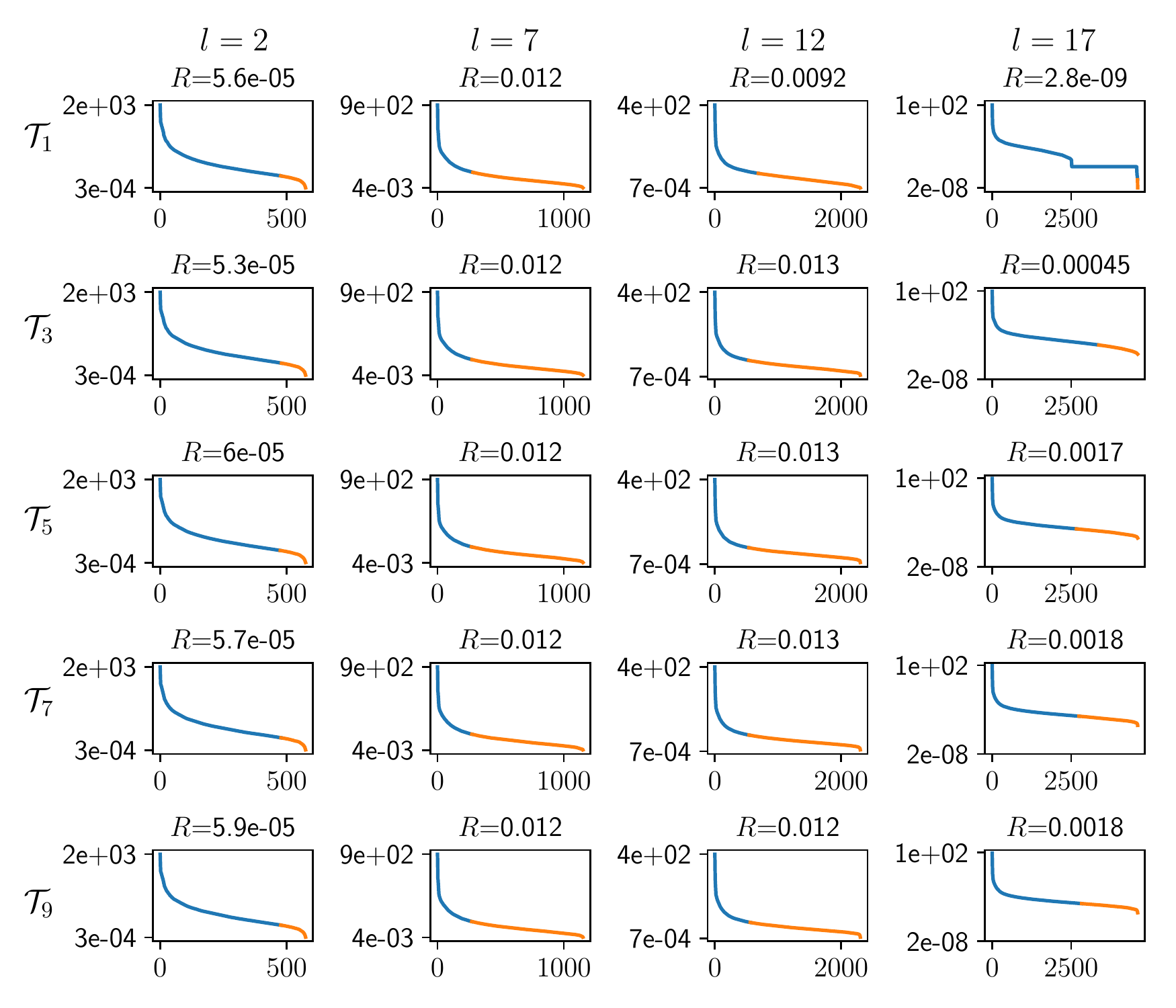}
	\caption{The illustration of proportion values of $R$ and singular values of uncentered covariance matrix at 2-th, 7-th, 12-th and 17-th linear layers of network trained on 10-split-CIFAR-100. Orange curves denote the singular values smaller than $10\lambda^l_{\text{min}}$.}
	\label{exp1eigen}
\end{figure}
%\subsubsection{10-split-CIFAR-100}
\textbf{10-split-CIFAR-100}. The comparative results on 10-split-CIFAR-100 are illustrated in \tabref{final10}, where the proposed Adam-NSCL achieves the highest ACC 73.77\% with competitive BWT -1.6\%.  The BWT values of GEM and A-GEM are better than Adam-NSCL, however, their ACC values are 49.48\% and 49.57\%, significantly lower than ours. EWC, LwF and GD-WILD achieve marginally worse ACC compared with Adam-NSCL, but the BWT values of LwF and GD-WILD are much lower. Both ACC and BWT values of  MAS, MUC-MAS and SI are much lower than ours. OWM has comparable BWT with our Adam-NSCL, but the ACC of OWM is 4.88\% lower than ours. Overall, our Adam-NSCL is the most preferable method among all these compared methods for continual learning.
%  It may be resulted in the approximate inversion of matrix and the approximate error is accumulated when incrementally computing the projection matrix. 

To justify the rationality of approximate null space, we show the curves of singular values in descending order and proportion values of $R$ defined in Eqn. \eqref{ratio} for the 2-th, 7-th, 12-th and 17-th layers of network in \figref{exp1eigen} on sequential tasks. As the results indicate, all proportion values of $U_2^l$ are smaller than 0.05, indicating that it is reasonable to take the range space of insignificant components $U_2^l$ as the approximate null space at the $l$-th layer $(l=1,\dots,L)$.

%which indicates that our method is the most preferable in comparison with other methods. This suggests that our method facilitates both the stability and plasticity of networks for continual learning where previous data is unavailable.
\begin{table}[!tbp]
	\begin{center}
		\scalebox{0.89}{
			\begin{tabular}{l|cc}
				\toprule
				\multirow{2}{5pt}{Method}
				&  \multicolumn{2}{|c}{20-split-CIFAR-100}   \\
				\cmidrule{2-3}
				&ACC (\%) & BWT(\%)  \\
				\midrule
				EWC \cite{kirkpatrick2017overcoming}   &71.66& -3.72  \\
				MAS  \cite{aljundi2018memory}  &63.84&   -6.29   \\
				MUC-MAS \cite{muc2020liu}  &67.22& -5.72\\
				SI \cite{osta2019learning}   &59.76& -8.62\\
				LwF \cite{li2017learning}  &74.38& -9.11\\
				InstAParam \cite{mitigating} &51.04&-4.92\\
				${}^*$GD-WILD \cite{lee2019overcoming}  & \textbf{77.16}& -14.85 \\%&$-$\\
				${}^*$GEM \cite{lopez2017gradient}  & 68.89& -1.2 \\
				${}^*$A-GEM \cite{chaudhry2018efficient} &61.91&   -6.88  \\%&$-$\\
				${}^*$MEGA \cite{guo2020improved}&64.98&-5.13\\
				OWM	\cite{zeng2019continual} &68.47& -3.37\\%&823\\
				\midrule
				Adam-NSCL  &\textbf{75.95}& -3.66 \\%&823\\ %58.82 -8.17
				\bottomrule
			\end{tabular}	
		}	
	\end{center}
	\caption{Comparisons of ACC and BWT for ResNet-18 sequentially trained on 20-split-CIFAR-100 using different methods.} % Methods required to store previous data are denoted by ${}^*$.
	\label{final20}
\end{table}

%\subsubsection{20-split-CIFAR-100}
\textbf{20-split-CIFAR-100}. The comparisons on 20-split-CIFAR-100 dataset are shown in \tabref{final20}. Our method achieves the second best ACC 75.95\%. Though GD-WILD achieves 1.21\% higher ACC than ours, the BWT of GD-WILD is 11.19\% lower than that of our Adam-NSCL. Furthermore, GD-WILD requires to save data of previous tasks and a large mount of external data. EWC, GEM and OWM achieve 4.29\%, 7.06\% and 7.48\% lower ACC values compared with our method. LwF has marginally lower ACC than ours, but its BWT value is significantly worse than ours. Other methods including MAS, MUC-MAS, SI and A-GEM fail to achieve comparable results as ours. Therefore, our Adam-NSCL outperforms the other compared methods for continual learning.

%Additionally, we verify it is reasonable to approximate the null space as done in \secref{sec:appns}, as shown in Appendix.

%\begin{table}[!htbp]
%	\begin{center}
%		\scalebox{0.85}{
%	\begin{tabular}{l|ccc}
%		\toprule
%		\multirow{2}{5pt}{Method}
%		&  \multicolumn{3}{|c}{25-split-TinyImageNet}   \\
%		\cmidrule{2-4}
%		&ACC (\%) & BWT(\%) &Memory Ratio(\%) \\
%		\hline
%		EWC \cite{kirkpatrick2017overcoming}   & 52.33  &-6.17 &2400\\
%		MAS  \cite{aljundi2018memory}  &47.96 &-7.04  &100 \\
%		MUC-MAS \cite{muc2020liu}  &41.60  &-3.73&400 \\
%		SI \cite{osta2019learning}   & 45.27 & -4.45&100\\
%		LwF \cite{li2017learning}  &56.57 & -11.19&$-$\\
%		${}^*$GD-WILD \cite{lee2019overcoming}  & 42.74  &-34.58 &$-$\\
%		${}^*$A-GEM \cite{chaudhry2018efficient} &53.32 &-7.68  &$-$\\
%		OWM	\cite{zeng2019continual} &49.98&-3.64&823\\
%		\hline
%		Ours   &58.28  &-6.05&823\\ %58.82 -8.17
%		\bottomrule
%	\end{tabular}	
%	}
%	\end{center}
%	\caption{The comparisons of ACC and BWT after ResNet-18 being sequentially trained on 25-split-TinyImageNet with different methods. } % Methods required to store previous data are denoted by ${}^*$.
%	\label{finaltiny}
%\end{table}

\begin{table}[!tbp]
	\begin{center}
		\scalebox{0.9}{
			\begin{tabular}{l|cc}
				\toprule
				\multirow{2}{5pt}{Method}
				&  \multicolumn{2}{|c}{25-split-TinyImageNet}   \\
				\cmidrule{2-3}
				&ACC (\%) & BWT(\%)  \\
				\midrule
				EWC \cite{kirkpatrick2017overcoming}   & 52.33  &-6.17\\% &2400\\
				MAS  \cite{aljundi2018memory}  &47.96 &-7.04  \\%&100 \\
				MUC-MAS \cite{muc2020liu}  &41.18  &-4.03\\%&400 \\
				SI \cite{osta2019learning}   & 45.27 & -4.45\\%&100\\
				LwF \cite{li2017learning}  &56.57 & -11.19\\%&$-$\\
				InstAParam \cite{mitigating} &34.64&-10.05\\
				${}^*$GD-WILD \cite{lee2019overcoming}  & 42.74  &-34.58 \\%&$-$\\
				${}^*$A-GEM \cite{chaudhry2018efficient} &53.32 &-7.68  \\%&$-$\\
				${}^*$MEGA \cite{guo2020improved} &57.12&-5.90\\
				OWM	\cite{zeng2019continual} &49.98&-3.64\\%&823\\
				\midrule
				Adam-NSCL   &\textbf{58.28}  &-6.05\\%&823\\ %58.82 -8.17
				\bottomrule
			\end{tabular}	
		}
	\end{center}
	\caption{Performance comparisons for ResNet-18 sequentially trained on 25-split-TinyImageNet using different methods.} % Methods required to store previous data are denoted by ${}^*$.
	\label{finaltiny}
\end{table}

\begin{figure}[!tbp]
	\centering
	\includegraphics[scale=0.89]{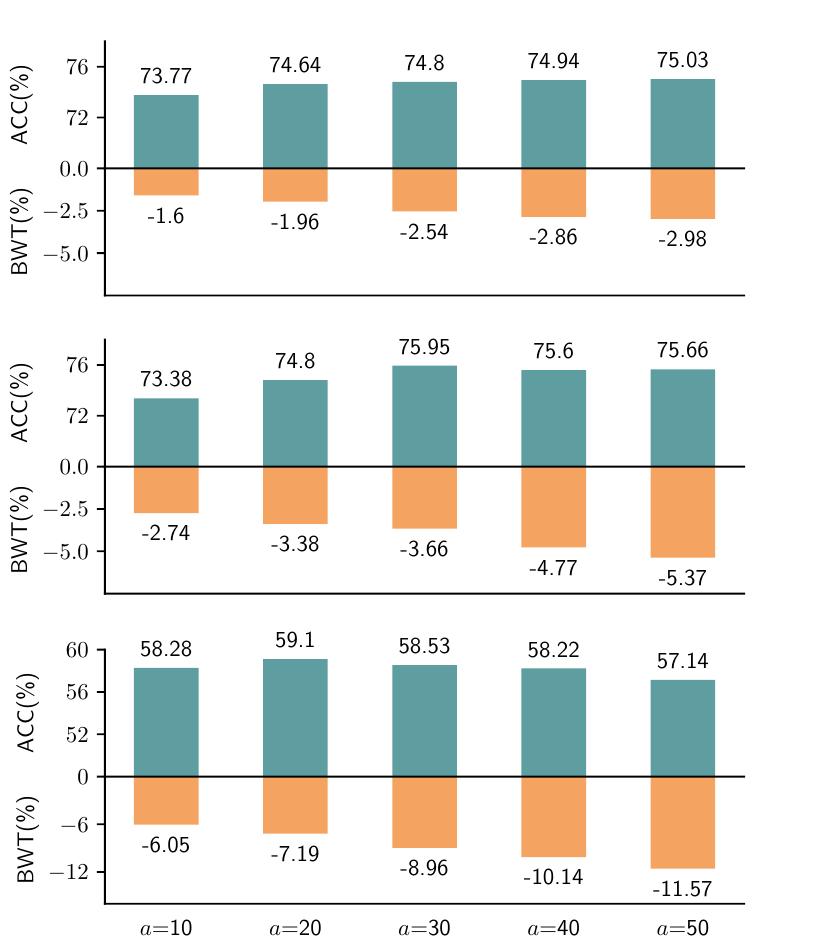}
	\caption{Stability and plasticity analysis. Top: 10-split-CIFAR-100. Middle:  20-split-CIFAR-100. Bottom: 25-split-TinyImageNet.}
	\label{ex:balance}
\end{figure}

\textbf{25-split-TinyImageNet}. As shown in \tabref{finaltiny}, on 25-split-TinyImageNet dataset, the proposed Adam-NSCL outperforms the other compared methods with the best ACC and competitive BWT values. Specifically, Adam-NSCL achieves the best ACC 58.28\% with comparable BWT -6.05\%. Though the BWT of Adam-NSCL is marginally lower than MUC-MAS, SI and OWM, these compared methods achieve 16.68\%, 13.01\% and 8.3\% lower ACC than ours. LwF achieves the second best ACC, but with much inferior BWT compared with our Adam-NSCL. With marginally lower BWT, EWC and MAS achieve 5.95\% and 10.32\% lower ACC than Adam-NSCL. 

%The rationality of the approximate null space is also verified, please refer to the Appendix.

We now discuss the difference between Adam-NSCL and OWM~\cite{zeng2019continual}. The main difference of Adam-NSCL and OWM is the way to find the null space as discussed in \secref{sec:related}. Computing the projection matrix in OWM relies on the approximate generalized inversion of feature matrix, and the approximate error can be accumulated when incrementally update the projection matrix. While in Adam-NSCL, the null space is specified by the uncentered feature covariance which can be incrementally computed without approximate error. Additionally, Adam-NSCL consistently performs better than OWM on 10-split-CIFAR-100 and 20-split-CIFAR-100, as shown in Tabs. \ref{final10} and \ref{final20}, where Adam-NSCL achieves 4.88\% and 7.48\% larger ACC and similar BWT in comparison with  OWM respectively. On 25-split-TinyImageNet, Adam-NSCL has significantly better ACC with comparable BWT than OWM, as shown in \tabref{finaltiny}.

\begin{figure}[!tb]
	\centering
	\includegraphics[scale=0.60]{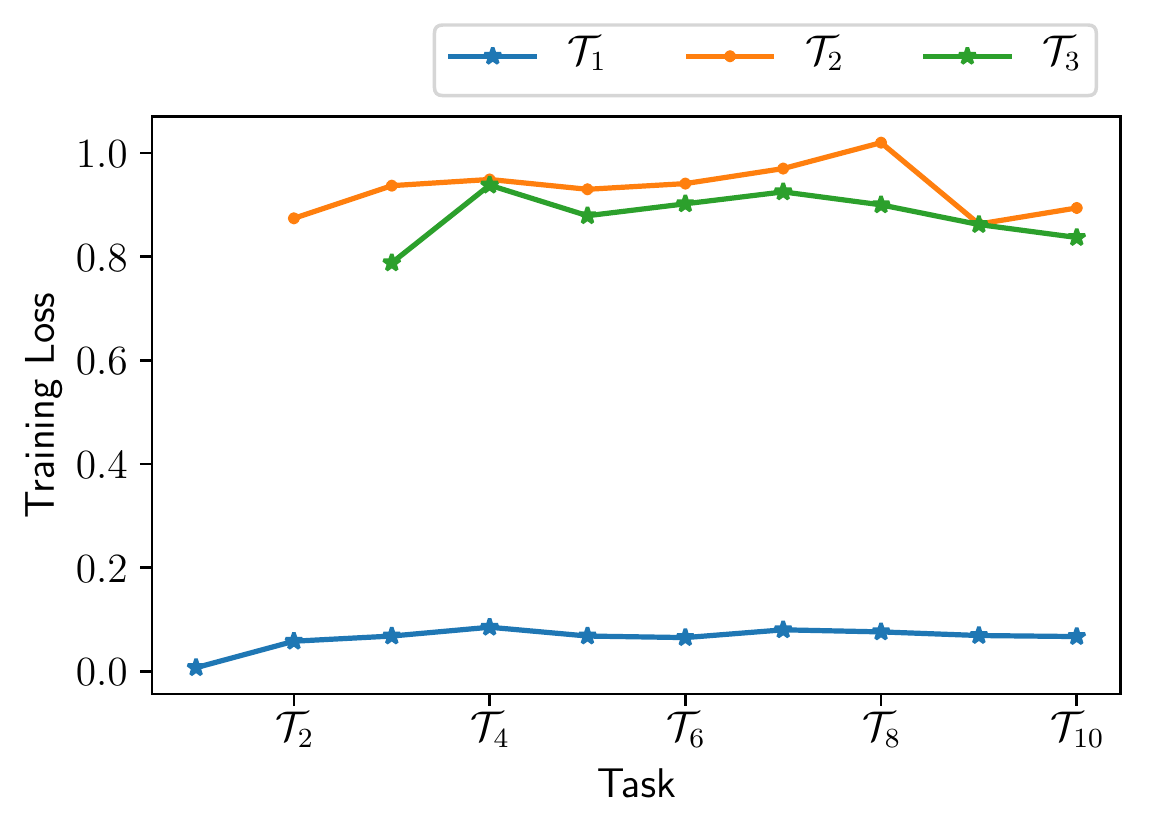}
	\caption{The curves of training losses of network on tasks $\mathcal{T}_1$, $\mathcal{T}_2$ and $\mathcal{T}_3$ when the network is trained on sequential tasks.}
	\label{exp:loss}
\end{figure}
\subsection{Model analysis} \label{sec:ana}
\textbf{Stability and plasticity analysis}. To study the balance of stability and plasticity, which is controlled by $a$, we compare the performance of our Adam-NSCL by varying $a=10,20,30,40,50$. According to \figref{ex:balance}, BWT becomes worse when $a$ is larger, suggesting that the network forgets more learned knowledge of previous tasks with lager $a$. Since ACC is affected by both of stability and plasticity, it increases first and then decreases with the increase of $a$ at the middle and bottom sub-figures of \figref{ex:balance}.

\textbf{Evolution of training loss}. To justify the proposed Adam-NSCL indeed guarantees the stability of network training on sequential tasks, we show the curves of training losses on the tasks $\mathcal{T}_1, \mathcal{T}_2, \mathcal{T}_3$ after learning new tasks in \figref{exp:loss} on 10-split-CIFAR-100. According to \figref{exp:loss}, the training losses of the network on previous tasks are retained after learning new tasks, verifying that Adam-NSCL, with Condition \ref{prop1} as basis, guarantees the stability of network.
%verifying the stability of network learning by our continual learning algorithm, guaranteed by Condition \ref{prop1}. 

\section{Conclusion}\label{conc}
In this paper, we address the \textit{plasticity-stability dilemma} for continual learning, constraining that the datasets of previous tasks are inaccessible. We propose two theoretical conditions to guarantee stability and plasticity for network parameter update when training networks on sequential tasks. Then we design a novel continual learning algorithm Adam-NSCL, which is based on Adam. The candidate parameter update generated by Adam is projected into the approximate null space of uncentered feature covariance matrix of previous tasks. Extensive experiments show that the proposed algorithm outperforms the compared methods for continual learning. In the future, we consider to further improve the approximation of null space and conduct theoretical analysis for our algorithm.

\textbf{Acknowledgment.} This work was supported by NSFC (U20B2075, 11690011, 11971373, U1811461, 12026605) and National Key R\&D Program 2018AAA0102201.

\pagebreak
%\widetext
\twocolumn[
\centering
\textbf{\huge Supplemental Materials}
\vspace{1cm}
]

We first introduce additional notations here. When feeding data $X_p$ from task $\mathcal{T}_p$ $(p\leq t)$ to $f$ with parameters $\mathbf{w}_{t,s}$, the input feature and output feature at the $l$-th linear layer are denoted as $X_{p,t,s}^l$ and $O_{p,t,s}^l$ respectively, then $$O^l_{{p,t,s}}=X^l_{{p,t,s}}  {w}^l_{{t,s}}, \ \ X^{l+1}_{{p,t,s}}=\sigma_l(O^l_{{p,t,s}})$$ with $X_{p,t,s}^1=X_p$. 
In addition, by denoting the learning rate as $\alpha$, we have $$w_{t,s}^l = w_{t,s-1}^l - \alpha \Delta w_{t,s-1}^l, \\ l=1,\dots,L.$$

\setcounter{lemma}{0}
\setcounter{prop}{1}
\section*{Appendix A}
In this appendix, we show the proof of Lemma \ref{lemma11} in the manuscript. Lemma \ref{lemma11} tells us that, when we train network on task $\mathcal{T}_t$, the network retains its training loss on data $X_p$ in the training process, if the network parameter update satisfies Eqn. \eqref{1} at each training step. We first recall Lemma \ref{lemma11} as follows, then give the proof.
\begin{lemma}\label{lemma11}
	Given the data $X_p$ from task $\mathcal{T}_p$,  and the network $f$ with $L$ linear layers is trained on task $\mathcal{T}_t$ ($t>p$). If network parameter update $\Delta w^l_{{t,s}}$ lies in the null space of $X^l_{{p,t-1}}$, \ie, 
	\begin{equation}\label{1}
		X^l_{p,t-1} \Delta w^l_{{t,s}} = 0,
	\end{equation}
	at each training step $s$, for the $l$-th layer of $f$ $(l=1,\dots,L)$,  we have $X^l_{{p,t}}=X^l_{p,t-1}$ and $f(X_p,\tilde{\mathbf{w}}_{t-1})=f(X_p,\tilde{\mathbf{w}}_{t})$.
\end{lemma}
\begin{proof}
	The proof is based on the recursive structure of network and iterative training process. We first prove that $X_{p,t,1}^l=X^l_{p,t-1}$ and $f(X_p, \mathbf{w}_{t,1}) = f(X_p, \tilde{\mathbf{w}}_{t-1})$ hold for $s=1$, and then illustrate that $X_{p,t,s}^l=X^l_{p,t-1}$ and $f(X_p, \mathbf{w}_{t,s}) = f(X_p, \tilde{\mathbf{w}}_{t-1})$ hold for each $s>1$, which suggests  that Lemma \ref{lemma1} holds.

	When $s=1$, considering that we initialize parameters $\mathbf{w}_{t,0}=\tilde{\mathbf{w}}_{t-1}$, we have
	\begin{equation}
		X^l_{{p,t,0}} =  X^l_{{p,t-1}}, \ \ O^l_{{p,t,0}} =  O^l_{{p,t-1}}.
	\end{equation}
	
	Therefore, at the first layer $(l=1)$ where $X_{p,t,1}^1=X_{p,t,0}^1=X^1_{{p,t-1}}$ (all of them equal to $X_p$ when $l=1$), 
	\begin{align}\label{layer}
		O_{p,t,1}^1 & = X_{p,t,1}^1 w_{t,1}^1 \notag \\
		&= X_{p,t,0}^1  (w_{t,0}^1 - \alpha \Delta w_{t,0}^1) \notag \\
		&= X_{p,t,0}^1  w_{t,0}^1 - \alpha X^1_{p,t-1} \Delta w_{t,0}^1 \notag \\
		&=X_{p,t,0}^1 w_{t,0}^1 \notag \\
		&=O_{p,t,0}^1,
	\end{align}
	where the fourth equation holds due to Eqn. \eqref{1}.
	Furthermore, we have 
	\begin{equation}\label{4}
		X_{p,t,1}^2=\sigma_1(O_{p,t,1}^1)=\sigma_1(O_{p,t,0}^1)=X_{p,t,0}^2=X_{p,t-1}^2,
	\end{equation}
	\ie, the input feature $X_{p,t,1}^2$ equals to $X_{p,t-1}^2$ at the second linear layer, based on which, we can recursively prove that 
	$$O_{p,t,1}^l=O_{p,t,0}^l=O_{p,t-1}^l$$
	and 
	$$X_{p,t,1}^l=X_{p,t,0}^l=X_{p,t-1}^l$$
	for $l=3, \dots, L$ by replacing $l=1$ with $l=2,\dots,L$ in Eqns. \eqref{layer} and \eqref{4}, then we have $f(X_p, \mathbf{w}_{t,1}) = f(X_p, $ $\tilde{\mathbf{w}}_{t-1})$.
	
	We now have proved that $X^l_{{p,t,s}} =  X^l_{{p,t-1}}$, $O^l_{{p,t,s}} =  O^l_{{p,t-1}}$ $(l=1,\dots,L)$ and $f(X_p, \mathbf{w}_{t,s}) = f(X_p, \tilde{\mathbf{w}}_{t-1})$ hold for $s=1$. Considering the iterative training process, we can prove that 
	\begin{equation*}
		X^l_{{p,t,s}} =  X^l_{{p,t-1}}, \ \ O^l_{{p,t,s}} =  O^l_{{p,t-1}} \ \ (l=1.\dots,L)
	\end{equation*}
	and $$f(X_p, \mathbf{w}_{t,s}) = f(X_p, \tilde{\mathbf{w}}_{t-1})$$
	hold for $s=2,...$, by repeating the above process with $s=2,...$.
	
	Finally, we have $X^l_{{p,t}}=X^l_{p,t-1}$ and $f(X_p,\tilde{\mathbf{w}}_{t-1})=f(X_p,\tilde{\mathbf{w}}_{t})$, since Lemma \ref{lemma1} holds for each $s\geq1$.
\end{proof}

\section*{Appendix B}
We first recall the Condition \ref{prop21} in the manuscript as follows, then prove that parameter update $\Delta\mathbf{w}_{t,s}$ satisfying condition \ref{prop21} is the descent direction, \ie, the training loss after updating parameters using $\Delta\mathbf{w}_{t,s}$ will decrease. 
\begin{prop}[plasticity]
	\label{prop21}
	Assume that the network $f$ is being trained on task $\mathcal{T}_t$, and $\mathbf{g}_{t,s}=\{g^1_{t,s},\dots,g^L_{t,s}\}$ denotes the parameter update generated by a gradient-descent training algorithm for training $f$ at training step $s$. $\langle\Delta \mathbf{w}_{t,s}, \mathbf{g}_{t,s}\rangle > 0$  should hold where $\langle \cdot, \cdot\rangle$ represents inner product.
\end{prop}
We now discuss the reason why $\Delta\mathbf{w}_{t,s}$ is the descent direction, if it satisfies condition \ref{prop21}. For clarity, we denote the loss for training network $f$ as $\mathcal{L}(\mathbf{w})$ which ignores the data term with no effect. The discussion can also be found in Lemma 2 of the lecture\footnote{\url{http://www.princeton.edu/~aaa/Public/Teaching/ORF363_COS323/F14/ORF363_COS323_F14_Lec8.pdf}}.

By denoting the learning rate as $\alpha$, and $h(\alpha) \triangleq \mathcal{L}(\mathbf{w}_{t,s}- \alpha \Delta\mathbf{w}_{t,s})$, according to Taylor's theorem, we have 
\begin{equation*}
	h(\alpha) = h(0) + \nabla_\alpha h(0) + o(\alpha),
\end{equation*}
\ie, 
\begin{align*}
	\mathcal{L}(\mathbf{w}_{t,s} - \alpha \Delta\mathbf{w}_{t,s}) = \mathcal{L}(\mathbf{w}_{t,s}) - \alpha  \langle\Delta \mathbf{w}_{t,s}, \mathbf{g}_{t,s}\rangle  + o(\alpha),
\end{align*}
where $\frac{|o(\alpha)|}{\alpha}\to0$ when $\alpha \to 0$. Therefore, there exists $\bar{\alpha}>0$ such that 
$$|o(\alpha)|< \alpha  |\langle\Delta \mathbf{w}_{t,s}, \mathbf{g}_{t,s}\rangle|, \ \  \forall \alpha \in (0, \bar{\alpha}).$$ Together with the condition $\langle\Delta \mathbf{w}_{t,s}, \mathbf{g}_{t,s}\rangle > 0$, we can conclude that $\mathcal{L}(\mathbf{w}_{t,s} - \alpha \Delta\mathbf{w}_{t,s})<\mathcal{L}(\mathbf{w}_{t,s})$ for all $ \alpha \in (0, \bar{\alpha})$. Therefore, parameter update $\Delta\mathbf{w}_{t,s}$ satisfying condition \ref{prop2} is the descent direction.
\section*{Appendix C}
Here, we give the proof of $\langle\Delta \mathbf{w}_{t,s}, \mathbf{g}_{t,s}\rangle \geq 0$  with $\Delta w^l_{t,s} = U_2^l(U_2^{l})^\top g_{t,s}^l$, which is claimed in Sec 4.1 of the manuscript.  The proof mainly utilizes the properties of Kronecker product \cite[Eqns. (2.10) and (2.13)]{graham2018kronecker}.
\begin{align}\label{inner}
	\langle\Delta \mathbf{w}_{t,s}, \mathbf{g}_{t,s}\rangle 
	&= \sum_{l=1}^L \langle U_2^l(U_2^{l})^\top g_{t,s}^l, g_{t,s}^l\rangle \notag \\
	&=\sum_{l=1}^L \text{vec}(U_2^l(U_2^{l})^\top g_{t,s}^lI)^\top \text{vec}(g_{t,s}^l)\notag \\
	&=\sum_{l=1}^L \text{vec}((U_2^{l})^\top g_{t,s}^l)^\top (I\otimes (U_2^{l})^\top )\text{vec}(g_{t,s}^l)\notag \\
	&=\sum _{l=1}^L \text{vec}((U_2^{l})^\top g_{t,s}^l)^\top\text{vec}((U_2^{l})^\top g_{t,s}^l) \notag \\
	&\geq0,
\end{align}
where vec$(\cdot)$ is the vectorization of $\cdot$, $I$ is the identity matrix and $\otimes$ is the Kronecker product. 

\section*{Appendix D}
We now discuss the difference between our algorithm and OWM \cite{zeng2019continual}  in details as follows. (1) We provide novel theoretical conditions for the stability and plasticity of network based on feature covariance. (2) The null space of ours is defined as the null space of feature covariance matrix which is easy to be accumulated after each task (refer to Q1 \& Alg. \textcolor{red}{2}). While the projection matrix in OWM is $\mathbf{P}_l=\mathbf{I}_l-\mathbf{A}_l(\mathbf{A}_l^\top \mathbf{A}_l+\beta_l \mathbf{I}_l)^{-1}\mathbf{A}_l^\top$ where $\mathbf{A}_l$ consists of all previous features of layer $l$. (3) With the coming of new tasks, our covariance matrix is incrementally updated without approximation error, while $\mathbf{P}_l$ of OWM is updated by recursive least square, where the approximation error of matrix inversion (because of the additionally introduced $\beta_l \mathbf{I}$) will be accumulated. (4) Our approach relies on a hyperparameter $a$ in line 14 of Alg. \textcolor{red}{2}, for approximating the null space of covariance, which can balance the stability and plasticity as discussed in lines 572-579 and Fig. \textcolor{red}{5}. It is easy to set the hyperparameter (line 614 and Figs. \textcolor{red}{4, 5}). But we find that it is hard to tune the hyperparameter $\beta_l$ in OWM for each layer to balance the approximation error and computational stability. (5) Experimental comparison with OWM on three benchmarks are shown in Tabs. \textcolor{red}{1-3}. The ACC of ours are 4.88\%, 7.48\% and 8.3\% higher than OWM with comparable BWT. Please refer to Q4 for comparison on ImageNet with deeper networks. We will clarify these differences by extending the discussions in Sect. \textcolor{red}{2}.

{\small
	\bibliographystyle{ieee_fullname}
	\bibliography{continuallearning}
}

\end{document}